\pdfoutput=1
\documentclass[letterpaper]{article}
\usepackage{aaai}
\usepackage{times}
\usepackage{helvet}
\usepackage{courier}
\usepackage{fixltx2e}

\usepackage{url}

\DeclareRobustCommand{\citep}[1]{\cite{#1}}
\DeclareRobustCommand{\citet}[1]{\citeauthor{#1}~\shortcite{#1}}
\usepackage{graphicx}
\usepackage{epstopdf} 
\usepackage{wrapfig}
\usepackage{capt-of} 
\usepackage{amsmath,amssymb}
\usepackage{bbm}

\usepackage{caption}
\usepackage{subcaption}

\usepackage{array,booktabs,dcolumn,calc}

\usepackage{url}

\usepackage[noabbrev,capitalize]{cleveref}
\crefformat{equation}{(#2#1#3)}
\crefrangeformat{equation}{(#3#1#4) to~(#5#2#6)}
\crefmultiformat{equation}{(#2#1#3)}%
{ and~(#2#1#3)}{, (#2#1#3)}{ and~(#2#1#3)}

\newcommand{\acro}[1]{\textsc{\MakeLowercase{#1}}}

\newcommand{\simiid}{\stackrel{iid}{\sim}}
\newcommand{\tp}{^\mathsf{T}}

\newcommand{\abs}[1]{\lvert #1 \rvert}
\newcommand{\Abs}[1]{\left\lvert #1 \right\rvert}
\newcommand{\norm}[1]{\lVert #1 \rVert}
\newcommand{\Norm}[1]{\left\lVert #1 \right\rVert}
\newcommand{\floor}[1]{\lfloor #1 \rfloor}

\newcommand{\E}{\mathbb{E}}

\newcommand{\ii}{\mathbbm{i}}

\newcommand{\N}{\mathcal{N}}
\newcommand{\R}{\mathbb{R}}

\newcommand{\Z}{\mathbb{Z}}

\newcommand{\where}{\mathrm{where}}

\newcommand{\ud}{\mathrm{d}}

\newcommand{\real}{\mathfrak{R}}
\newcommand{\imag}{\mathfrak{I}}

\DeclareMathOperator{\diag}{diag}

\DeclareMathOperator{\Unif}{Unif}

\DeclareMathOperator{\JS}{JS}
\DeclareMathOperator{\TV}{TV}

\DeclareMathOperator{\uH}{H}

\begin{document}
\title{Linear-time Learning on Distributions with Approximate Kernel Embeddings}
\renewcommand{\thefootnote}{\fnsymbol{footnote}}
\author{
Danica J.\ Sutherland\footnotemark[1]
\and Junier B.\ Oliva\footnotemark[1]
\and Barnab\'as P\'oczos
\and Jeff Schneider
\\
Carnegie Mellon University \\
\{dsutherl,joliva,bapoczos,schneide\}@cs.cmu.edu
}
\nocopyright
\maketitle
\begin{abstract}
\begin{quote}
Many interesting machine learning problems are best posed by considering instances that are distributions, or sample sets drawn from distributions.  
Previous work devoted to machine learning tasks with distributional inputs has done so through pairwise kernel evaluations between pdfs (or sample sets).  While such an approach is fine for smaller datasets, the computation of an $N \times N$ Gram matrix is prohibitive in large datasets.
Recent scalable estimators that work over pdfs have done so only with kernels that use Euclidean metrics, like the $L_2$ distance.
However, there are a myriad of other useful metrics available, such as total variation, Hellinger distance, and the Jensen-Shannon divergence.
This work develops the first random features for pdfs whose dot product approximates kernels using these non-Euclidean metrics,
allowing estimators using such kernels to scale to large datasets by working in a primal space, without computing large Gram matrices.
We provide an analysis of the approximation error in using our proposed random features
and show empirically the quality of our approximation both in estimating a Gram matrix and in solving learning tasks in real-world and synthetic data.
\end{quote}
\end{abstract}

\section{Introduction}

\footnotetext[1]{These two authors contributed equally.}
\renewcommand{\thefootnote}{\arabic{footnote}}

As machine learning matures, focus has shifted towards datasets with richer, more complex instances.
For example, a great deal of effort has been devoted to learning functions on vectors of a large fixed dimension.
While complex static vector instances are useful in a myriad of applications, many machine learning problems are more naturally posed by considering 
instances that are distributions, or sets drawn from distributions. 
Political scientists can learn a function from community demographics to vote percentages to understand who supports a candidate \citep{seth:ecological}.
The mass of dark matter halos can be inferred from the velocity of galaxies in a cluster \citep{sdm:apj}.
Expensive expectation propagation messages can be sped up by learning a ``just-in-time'' regression model \citep{jitkrittum2015kernel}.
All of these applications are aided by working directly over sets drawn from the distribution of interest, rather than having to develop a per-problem ad-hoc set of summary statistics.

Distributions are inherently infinite-dimensional objects, since in general they require an infinite number of parameters for their exact representation. Hence, it is not immediate how to extend traditional finite vector technique machine learning techniques to distributional instances.
However, recent work has provided various approaches for dealing with distributional data in a nonparametric fashion.
For example, regression from distributional covariates to real or distributional responses is possible via kernel smoothing \citep{poczos2013distribution,oliva2013distribution},
and many learning tasks can be solved with \acro{RKHS} approaches \citep{muandet2012learning,poczos2012nonparametric}.
A major shortcoming of both approaches is that they require computing $N$ kernel evaluations per prediction, where $N$ is the number of training instances in a dataset.
Often, this implies that one must compute a $N\times N$ Gram matrix of pairwise kernel evaluations.
Such approaches fail to scale to datasets where the number of instances $N$ is very large.
Another shortcoming of these approaches is that they are often based on Euclidean metrics, either working over a linear kernel, or one based on the $L_2$ distance over distributions.
While such kernels are useful in certain applications, better performance can sometimes be obtained by considering non-Euclidean based kernels.
To this end, \citet{poczos2012nonparametric} use a kernel based on R\'{e}nyi divergences; however, this kernel is not positive semi-definite (\acro{psd}),
leading to even higher computational cost and other practical issues.

\begin{figure}[b]
  \centering
  \includegraphics[width=.45\textwidth]{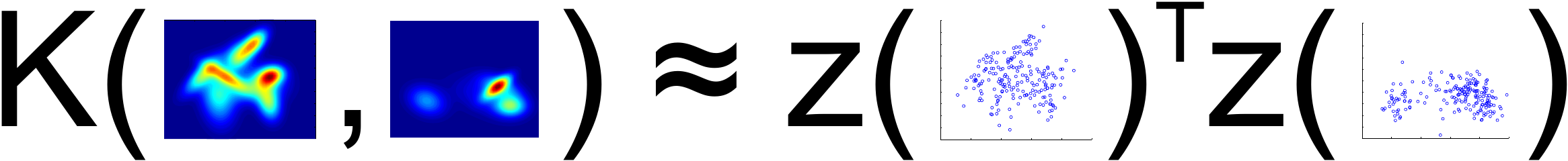}
  \caption{\label{fig:kernImag}We approximate kernels between densities $p_i,p_j$ with
  random features of sample sets $\chi_i\simiid p_i, \chi_j\simiid p_j$.}
\end{figure}

This work addresses these major shortcomings by developing an embedding of random features for distributions. The dot product of the random features for two distributions will approximate kernels based on various distances between densities (see \Cref{fig:kernImag}).
With this technique, we can approximate kernels based on total variation, Hellinger, and Jensen-Shannon divergences, among others.
Since there is then no need to compute a Gram matrix,
one will be able to use these kernels while still scaling to datasets with a large number of instances
using primal-space techniques.
We provide an approximation bound for the embeddings, and demonstrate the efficacy of the embeddings on both real-world and synthetic data.
To the best of our knowledge, this work provides the first non-discretized embedding for non-$L_2$ kernels for probability density functions.

\section{Related Work}

The two main lines of relevant research are
the development of kernels on probability distributions
and explicit approximate embeddings for scalable kernel learning.

\paragraph{Learning on distributions}
% We review some of the major methods for learning on distributional inputs here.
In computer vision, the popular ``bag of words'' model \citep{textons} represents a distribution by quantizing it onto codewords (usually by running $k$-means on all, or many, of the input points from all sets), then compares those histograms with some kernel (often exponentiated $\chi^2$).

Another approach estimates a distance between distributions,
often the $L_2$ distance or Kullback-Leibler (\acro{KL}) divergence,
parametrically \citep{Jaakkola98,Moreno2003,Jebara04ppk}
or nonparametrically \citep{Sricharan2012,renyi-friends}.
The distance can then be used in
kernel smoothing \cite{poczos2013distribution,oliva2013distribution}
or Mercer kernels \cite{Moreno2003,Kondor2003,Jebara04ppk,poczos2012nonparametric}.

These approaches can be powerful, but usually require computing an $N \times N$ matrix of kernel evaluations, which can be infeasible for large datasets.
Using these distances in Mercer kernels faces an additional challenge,
which is that the estimated Gram matrix may not be \acro{PSD},
due to estimation error or because some divergences in fact do not induce a \acro{PSD} kernel.
In general this must be remedied by altering the Gram matrix a ``nearby'' \acro{PSD} one.
Typical approaches involve eigendecomposing the Gram matrix,
which usually costs $O(N^3)$ computation and also presents challenges for traditional inductive learning, where the test points are not known at training time \cite{Chen2009}.

One way to alleviate the scaling problem is the Nystr\"om extension \citep{Williams2001},
in which some columns of the Gram matrix are used to estimate the remainder.
In practice,
one frequently must compute many columns,
and methods to make the result \acro{PSD} are known only for mildly-indefinite kernels \citep{Belongie2002}.

Another approach is to represent a distribution by its mean \acro{RKHS} embedding under some kernel $k$.
The \acro{RKHS} inner product is known as the \emph{mean map kernel} (\acro{MMK}),
and the distance the \emph{maximum mean discrepancy} (\acro{MMD})
\citep{Gretton2009,muandet2012learning,szabo2014two}.
When $k$ is the common \acro{RBF} kernel,
the \acro{MMK} estimate is proportional to an $L_2$ inner product between Gaussian kernel density estimates.

\paragraph{Approximate embeddings}
Recent interest in approximate kernel embeddings was spurred by 
the ``random kitchen sink'' (\acro{rks}) embedding
\citep{rahimi2007random},
which approximates shift-invariant kernels $K$ on $\R^\ell$
by sampling their Fourier transform.

A related line of work considers additive kernels,
of the form $K(x, y) = \sum_{j=1}^\ell \kappa(x_j, y_j)$,
usually defined on ${\R^\ell_{\ge 0}}$ (e.g.\ histograms).
\citet{Maji2009} construct an embedding for the intersection kernel $\sum_{j=1}^\ell \min(x_j, y_j)$ via step functions.
\citet{Vedaldi-cvpr} consider any homogeneous $\kappa$, so that $\kappa(tx, ty) = t \, \kappa(x, y)$, which also allows them to embed histogram kernels such as the additive $\chi^2$ kernel and Jensen-Shannon divergence.
Their embedding uses the same fundamental result of \citet{Fuglede2005} as ours; we expand to the continuous rather than the discrete case.
\citet{Vempati2010} later apply \acro{rks} embeddings to obtain generalized \acro{rbf} kernels \cref{eq:InfKern}.

For embeddings of kernels on input spaces other than $\R^\ell$,
the \acro{RKS} embedding extends naturally to locally compact abelian groups \citep{Li2010}.
\citet{Oliva2014} embedded an estimate of the $L_2$ distance between continuous densities via orthonormal basis functions.
An embedding for the base kernel $k$ also gives a simple embedding for the mean map kernel \citep{seth:ecological,jitkrittum2015kernel,lopez2015towards,rff-error}.

\section{Embedding Information Theoretic Kernels}
For a broad class of distributional distances $d$,
including many common and useful information theoretic divergences,
we consider generalized \acro{rbf} kernels of the form
\begin{align}
  K(p, q) = \exp\left( - \frac{1}{2\sigma^2} d^2(p, q) \right)
  \label{eq:InfKern}
.\end{align}
We will construct features $z(A(\cdot))$ such that
$K(p, q) \approx z(A(p))\tp z(A(q))$
as follows:
\begin{enumerate}
\item
  We define a random function $\psi$
  such that $d(p, q) \approx \norm{\psi(p) - \psi(q)}$,
  where $\psi(p)$ is a function from $[0, 1]^\ell$ to $\R^{2 M}$.
  Thus the metric space of densities with distance $d$ is approximately embedded into the metric space of $2M$-dimensional $L_2$ functions.
\item
  We use orthonormal basis functions to approximately embed smooth $L_2$ functions
  into finite vectors in $\R^{\abs V}$.
  Combined with the previous step,
  we obtain features $A(p) \in \R^{2 M \abs V}$ such that $d$ is approximated by Euclidean distances between the $A$ features.
\item 
  We use the \acro{rks} embedding $z(\cdot)$ so that inner products between $z(A(\cdot))$ features, in $\R^D$, approximate $K(p, q)$.
\end{enumerate}
We can thus approximate the powerful kernel $K$ without needing to compute an expensive $N \times N$ Gram matrix.

\subsection{Homogeneous Density Distances (\textsc{hdd}s)}
We consider kernels based on metrics which we term homogeneous density distances (\acro{HDD}s):
\begin{align}
  d^2(p,q) = \int_{[0,1]^\ell} \kappa(p(x),q(x)) \,\ud x
  \label{eq:hdd-def}
,\end{align}
where $\kappa(x,y):\R_+ \times \R_+ \to \R_+$ is a negative-type kernel,
i.e.\ a squared Hilbertian metric,
and $\kappa(tx, ty) = t \kappa(x, y)$ for all $t > 0$.
\Cref{tab:hdds} shows a few important instances.
Note that we assume the support of the distributions is contained within $[0, 1]^\ell$.

\begin{table}[ht]
\centering
\begin{tabular}{l{c}{c}}
Name      & $\kappa(p(x),q(x))$ & $\ud\mu(\lambda)$ \\
\hline
% Jensen-Shannon
$\JS$
% & $
%     \frac{p(x)}{2}\log\left(\frac{2p(x)}{p(x)+q(x)}\right) +
%     \frac{q(x)}{2}\log\left(\frac{2q(x)}{p(x)+q(x)}\right)
%   $
& $\sum_{r \in \{p, q\}} \tfrac12 r(x) \log\left( \frac{2 r(x)}{p(x)+q(x)} \right)$
& $\frac{\ud \lambda}{\cosh(\pi\lambda)(1+\lambda^2)}$
\\
% Squared Hellinger
$\uH^2$
& % ${p(x)+q(y)} - \tfrac12 \left( {p(x)^\frac{1}{2}+q(y)^\frac{1}{2}}  \right)^2 $
  $\tfrac12 \left( \sqrt{p(x)} - \sqrt{q(x)} \right)^2$
& $\tfrac12 \,\delta(\lambda=0) \,\ud\lambda$
\\
% Total Variation
$\TV$
& % $\max\left(p(x),q(x)\right) - \frac{p(x)+q(y)}{2}$
  $\abs{p(x) - q(x)}$
& $\frac{2}{\pi}\frac{1}{1+4\lambda^2} \,\ud \lambda$ \\
\end{tabular}
\caption{Squared \acro{HDD}s.
$\JS$ is Jensen-Shannon divergence; $\uH$ is Hellinger distance; $\TV$ is total variation distance.}
\label{tab:hdds}
\end{table}

We then use these distances in a generalized \acro{RBF} kernel \cref{eq:InfKern}.
$d$ is a Hilbertian metric \citep{Fuglede2005},
so $K$ is positive definite \citep{haasdonk:dist-sub}.
Note we use the $\sqrt{\TV}$ metric, even though $\TV$ is itself a metric.

\subsection{Embedding \textsc{hdd}s into $L_2$}
\citet{Fuglede2005} shows that $\kappa$ corresponds to a bounded measure $\mu(\lambda)$, as in \cref{tab:hdds}, with
\begin{equation}
\kappa(x, y) = \int_{\R_{\ge 0}} \abs{x^{\tfrac12 + \ii \lambda} - y^{\tfrac12 + \ii \lambda}}^2 \,\ud\mu(\lambda)
.\end{equation}
Let
$Z = \mu(R_{\ge 0})$
and $c_\lambda = (-\tfrac12 + \ii\lambda)/(\tfrac12 + \ii\lambda)$;
then 
\begin{equation*}
\kappa(x, y) = \E_{\lambda \sim \frac{\mu}{Z} } \abs{g_\lambda(x) - g_\lambda(y)}^2
\end{equation*}
where
$g_\lambda(x) = \sqrt{Z} c_\lambda (x^{\frac12 + i \lambda} - 1)$.

We can approximate the expectation with an empirical mean.
Let $\lambda_j\simiid\frac{\mu}{Z}$ for $j\in \{1,\ldots,M\}$;
then
\begin{gather*}
  \kappa(x,y)
  \approx \frac{1}{M} \sum_{j=1}^M \abs{g_{\lambda_j}(x) - g_{\lambda_j}(y)}^2
.\end{gather*}
Hence, using $\real, \imag$ to denote the real and imaginary parts:
\begin{align}
d^2(p, q)
&= \int_{[0,1]^\ell} \kappa(p(x),q(x))\, \ud x \nonumber
\\&= \int_{[0,1]^\ell} \E_{\lambda \sim \frac{\mu}{Z}} \abs{g_\lambda(p(x)) - g_\lambda(q(x))}^2\, \ud x \nonumber\\
&\approx \frac{1}{M}\sum_{j=1}^M \int_{[0,1]^\ell}
\big(
\left(\real(g_{\lambda_j}(p(x))) - \real(g_{\lambda_j}(q(x))) \right)^2
\nonumber\\
&\qquad\qquad
+\left(\imag(g_{\lambda_j}(p(x))) - \imag(g_{\lambda_j}(q(x))) \right)^2 \big) \ud x \nonumber\\
&= \norm{\psi(p)-\psi(q)}^2,
\end{align}
where $[\psi(p)](x)$ is given by
\[
\frac{1}{\sqrt{M}} \big(
        p^R_{\lambda_1}(x), \ldots, p^R_{\lambda_M}(x), 
        p^I_{\lambda_1}(x),\ldots,p^I_{\lambda_M}(x)
     \big)
,\]
defining
$p^R_{\lambda_j}(x) = \real(g_{\lambda_j}(p(x)))$,
$p^I_{\lambda_j}(x) = \imag(g_{\lambda_j}(p(x)))$.
Hence, the \acro{HDD} between densities $p$ and $q$ is approximately the $L_2$ distance from $\psi(p)$ to $\psi(q)$,
where $\psi$ maps a function $f:[0,1]^\ell \mapsto \R$ to a vector-valued function $\psi(f):[0,1]^\ell \mapsto \R^{2M}$ of $\lambda$ functions.
$M$ can typically be quite small, since the kernel it approximates is one-dimensional.

\subsection{Finite Embeddings of $L_2$}
If densities $p$ and $q$ are smooth,
then the $L_2$ metric between the $p_\lambda$ and $q_\lambda$ functions
may be well approximated using projections to basis functions.
Suppose that $\{\varphi_i\}_{i\in\Z}$ is an orthonormal basis for $L_2([0,1])$;
then we can construct an orthonormal basis for $L_2([0, 1]^\ell)$ by the tensor product:
\begin{gather*}
    \{\varphi_\alpha\}_{\alpha\in\Z^\ell}
    \qquad \where \quad
    \varphi_\alpha(x) = \prod_{i=1}^\ell \varphi_{\alpha_i}(x_i),\ x\in [0,1]^\ell,\\
    \forall f \in L_2([0,1]^\ell),\   f(x) = \sum_{\alpha\in\Z^\ell} a_\alpha(f) \, \varphi_\alpha(x)
\end{gather*}
and $a_\alpha(f) = \langle\varphi_\alpha, f\rangle = \int_{[0,1]^\ell} \varphi_\alpha(t) \, f(t)\, \ud t \in \R$.
Let $V \subset \Z^\ell$ be an appropriately chosen finite set of indices.
If $f, f' \in L_2([0,1]^\ell)$ are smooth and $\vec{a}(f) = (a_{\alpha_1}(f), \ldots, a_{\alpha_{\abs V}}(f))$, then $\norm{f-f'}^2 \approx \norm{\vec{a}(f)-\vec{a}(f')}^2$.
Thus we can approximate $d^2$ as the squared distance between finite vectors:
\begin{align}
    d^2(p,q)
     &\approx \norm{\psi(p)-\psi(q)}^2 \nonumber \\
     &\approx \frac1M \sum_{j=1}^M \norm{
        \vec{a}(p^R_{\lambda_j})
      - \vec{a}(q^R_{\lambda_j})
     }^2
  + \norm{
        \vec{a}(p^I_{\lambda_j})
      - \vec{a}(q^I_{\lambda_j})
     }^2 \nonumber \\
     &= \norm{A(p) - A(q)}^2 \label{eq:Anorms}
\end{align}
where $A : L_2([0, 1]^\ell) \to \R^{2M\abs{V}}$ concatenates the $\vec a$ features for each $\lambda$ function.
That is, $A(p)$ is given by
\begin{equation}
    \frac{1}{\sqrt M} \left(
        \vec{a}(p^R_{\lambda_1}), \ldots, \vec{a}(p^R_{\lambda_M}),  
        \vec{a}(p^I_{\lambda_1}), \ldots, \vec{a}(p^I_{\lambda_M})
    \right).
    \label{eq:Adef}
\end{equation}
We will discuss how to estimate $\vec{a}(p_\lambda^R)$, $\vec{a}(p_\lambda^I)$ shortly.

\subsection{Embedding \textsc{rbf} Kernels into $\R^D$}

The $A$ features approximate the \acro{HDD} \eqref{eq:hdd-def} in $\R^{2 M \Abs{V}}$;
thus applying the \acro{RKS} embedding \citep{rahimi2007random} to the $A$ features will approximate our generalized \acro{RBF} kernel \eqref{eq:InfKern}.
The \acro{RKS} embedding 
is\footnote{There are two versions of the embedding in common use, but this one is preferred \citep{rff-error}.} $z : \R^m \to \R^D$ such that for fixed
$\{ \omega_i \}_{i=1}^{D/2} \simiid \N(0,\sigma^{-2}I_m)$
and for each $x,y \in \R^m$:
 \begin{align}
    &z(x)\tp z(y) \approx \exp\left(- \tfrac{1}{2\sigma^2} \Norm{x-y}^2\right),\ \where \notag\\
    &z(x) =
 \sqrt{\tfrac{2}{D}}
\left( \sin(\omega_1\tp x),\cos(\omega_1\tp x), \ldots \right)
 \label{eq:rks_feats}
.\end{align}

Thus we can approximate the \acro{HDD} kernel \eqref{eq:InfKern} as:
\begin{align}
K(p,q)
&= \exp\left(-\frac{1}{2\sigma^2} d^2(p,q) \right) \notag \\
&\approx \exp\left(-\frac{1}{2\sigma^2} \norm{A(p)-A(q)}^2 \right) \notag \\
&\approx z(A(p))\tp z(A(q))
\label{eq:fullapprox}.
\end{align}

\subsection{Finite Sample Estimates}
Our final approximation for \acro{HDD} kernels \eqref{eq:fullapprox} depends on integrals of densities $p$ and $q$.
In practice, we are unlikely to directly observe an input density,
but even given a pdf $p$, the integrals that make up the elements of $A(p)$ are not readily computable. 
We thus first estimate the density as $\hat{p}$, e.g.\ with kernel density estimation (\acro{kde}),
and estimate $A(p)$ as $A(\hat{p})$.
Recall that the elements of $A(\hat{p})$ are:
\begin{align}
    a_\alpha(\hat{p}_{\lambda_j}^S)
    =
    \int_{[0,1]^\ell} \varphi_\alpha(t) \, \hat{p}_{\lambda_j}^S(t) \,\ud t
    \label{eq:ApInt}
\end{align}
where $j \in \{1, \ldots, M\}, S \in \{R, I\}, \alpha \in V$.
In lower dimensions, we can approximate \eqref{eq:ApInt} with simple Monte Carlo numerical integration. Choosing $\{ u_i \}_{i=1}^{n_e} \simiid \Unif([0,1]^\ell)$:
\begin{align}
    \hat{a}_\alpha(\hat{p}_{\lambda_j}^S)
    =
    \frac{1}{n_e} \sum_{i=1}^{n_e} \varphi_\alpha(u_i) \,\hat{p}_{\lambda_j}^S(u_i)
    \label{eq:ApNumInt}
,\end{align}
obtaining $\hat{A}(\hat p)$.
We note that in high dimensions, one may use any high-dimensional density estimation scheme (e.g.\ \citeauthor{lafferty:sparse-density}~\citeyear{lafferty:sparse-density})
and estimate \eqref{eq:ApInt} with \acro{MCMC} techniques (e.g.\ \citeauthor{nuts}~\citeyear{nuts}).

\subsection{Summary and Complexity}
The algorithm for computing features $\{z(A(p_i))\}_{i=1}^N$ for a set of distributions $\{p_i\}_{i=1}^N$, given sample sets $\{\chi_i\}_{i=1}^N$ where $\chi_i = \{X_j^{(i)}\in [0,1]^\ell\}_{j=1}^{n_i} \simiid p_i$, is thus:
\begin{enumerate}
  \item Draw
      $M$ scalars $\lambda_j \simiid \frac{\mu}{Z}$
      and $D/2$ vectors $\omega_r \simiid \N(0, \sigma^{-2}I_{2 M \Abs{V}})$,
      in $O(M \Abs{V} D)$ time.
  \item For each of the $N$ input distributions $i$:
  \begin{enumerate}
  \item Compute a kernel density estimate from $\chi_i$, $\hat{p}_i(u_j)$ for each $u_j$ in \eqref{eq:ApNumInt},
    in $O(n_i n_e)$ time.
  \item Compute $\hat{A}(\hat{p}_i)$ using a numerical integration estimate as in \eqref{eq:ApNumInt},
    in $O( M \Abs{V} n_e )$ time.
  \item Get the \acro{rks} features, $z(\hat{A}(\hat{p}_i))$,
    in $O( M \Abs{V} D )$ time.
\end{enumerate}
\end{enumerate}
Supposing each $n_i \asymp n$,
this process takes a total of
$O\left( N n n_e + N M \Abs{V} n_e + N M \Abs{V} D \right)$ time.
Taking $\Abs{V}$ to be asymptotically $O(n)$,
$n_e = O(D)$,
and $M = O(1)$ for simplicity,
this is $O(N n D)$ time,
compared to 
$O(N^2 n \log n + N^3)$ for the methods of \citet{poczos2012nonparametric}
and $O(N^2 n^2)$ for \citet{muandet2012learning}.

\section{Theory}
\label{sec:theory}

We bound
$\Pr\left(\Abs{K(p, q) - z(\hat{A}(\hat{p}))\tp z(\hat{A}(\hat{q}))} \ge \varepsilon\right)$
for two fixed densities $p$ and $q$
by considering each source of error:
kernel density estimation $(\varepsilon_\text{KDE})$;
approximating $\mu(\lambda)$ with $M$ samples ($\varepsilon_\lambda$);
truncating the tails of the projection coefficients ($\varepsilon_\text{tail}$);
Monte Carlo integration ($\varepsilon_\text{int}$);
and the $\acro{rks}$ embedding ($\varepsilon_\text{RKS}$). 

We need some smoothness assumptions on $p$ and $q$:
that they are members of a periodic H\"older class $\Sigma_\text{per}(\beta, L_\beta)$,
that they are bounded below by $\rho_\ast$ and above by $\rho^\ast$,
and that their kernel density estimates are
in $\Sigma_\text{per}(\hat\gamma, \widehat L)$ with probability at least $1 - \delta$.
We use a suitable form of kernel density estimation,
to obtain a uniform error bound with a rate based on the function $C^{-1}$ \citep{Gine2002}.
We use the Fourier basis and choose $V = \{ \alpha \in \Z^\ell \mid \sum_{j=1}^\ell \abs{\alpha_j}^{2 s} \le t \}$ for parameters $0 < s < \hat\gamma$, $t > 0$.

Then,
for any
$\varepsilon_\text{RKS} + \frac{1}{\sigma_k \sqrt{e}} \left( \varepsilon_\text{KDE} + \varepsilon_\lambda + \varepsilon_\text{tail} + \varepsilon_\text{int} \right) \le \varepsilon$,
the probability of the error exceeding $\varepsilon$ is at most:
\begin{multline*}
      2 \exp\left( - D \varepsilon_\text{RKS}^2 \right)
    + 2 \exp \left( - M \varepsilon_{\lambda}^4 / (8 Z^2) \right)
    + \delta
\\  + 2 C^{-1}\left( \frac{\varepsilon^4_\text{KDE} n^{2 \beta / (2 \beta + \ell)}}{4 \log n} \right)
    + 2 M \left(
    1
    - \mu\big( [0, u_\text{tail}) \big)
    \right)
\\  + 8 M \,\abs{V} \exp\left( - \tfrac12 n_e \left( \frac{\sqrt{1 + \varepsilon_\text{int}^2 / (8 \,\abs{V}\, Z)} - 1}{\sqrt{\rho^*} + 1} \right)^2 \right)
\end{multline*}
where $u_\text{tail} = \sqrt{ \max\left(0, 
        \frac{\rho_* t }{8 M \ell \widehat{L}^2}
        \frac{4^{\hat\gamma} - 4^s}{4^{\hat\gamma}} \varepsilon_\text{tail}^2
        - \frac14 \right)}$.

The bound decreases
when the function is smoother (larger $\beta$, $\hat\gamma$; smaller $\widehat L$)
or lower-dimensional ($\ell$),
or when we observe more samples ($n$).
Using more projection coefficients (higher $t$ or smaller $s$, giving higher $\abs V$) improves the approximation but makes numerical integration more difficult.
Likewise, taking more samples from $\mu$ (higher $M$) improves that approximation,
but increases the number of functions to be approximated and numerically integrated.

For the proof and further details, see the appendix.

\section{Numerical Experiments}
\label{sec:experiments}
Throughout these experiments we use $M=5$, $\abs{V}=10^\ell$ (selected as rules of thumb; larger values did not improve performance),
and use a validation set (10\% of the training set)
to choose bandwidths for \acro{KDE} and the \acro{RBF} kernel as well as model regularization parameters.
Except in the scene classification experiments, 
the histogram methods used 10 bins per dimension;
performance with other values was not better.
The \acro{kl} estimator used the fourth nearest neighbor.

We evaluate \acro{rbf} kernels based on various distances.
First, we try our {\small \textsf{JS}}, {\small \textsf{Hellinger}}, and {\small \textsf{TV}} embeddings.
We compare to $L_2$ kernels
as in \citet{Oliva2014}:
$    \exp\left( -\frac{1}{2\sigma^2} \norm{p-q}_2^2 \right)
    \approx
    z(\vec{a}(\hat{p}))\tp z(\vec{a}(\hat{q}))$
({\small \textsf{L2}}).
We also try the \acro{MMD} distance \citep{muandet2012learning} 
with approximate kernel embeddings:
    $\exp\left( - \tfrac{1}{2 \sigma^2} \widehat{\mathrm{MMD}}({p}, {q}) \right)
    \approx z\left(\bar{z}(\hat{p}) \right)\tp z\left(\bar{z}(\hat{q}) \right)$, 
    where $\bar z$ is the mean embedding
    $\bar{z}(\hat{p}) = \frac{1}{n} \sum_{i=1}^n z(X_i)$
({\small \textsf{MMD}}).
We further compare to \acro{rks} with histogram \acro{JS} embeddings \citep{Vempati2010}
({\small \textsf{Hist JS}});
we also tried $\chi^2$ embeddings, but their performance was quite similar.
We finally try the full Gram matrix approach of \citet{poczos2012nonparametric}
with the \acro{KL} estimator of \citet{wang2009} in an \acro{rbf} kernel
({\small \textsf{KL}}),
as did \citet{sdm:apj}.

\subsection{Gram Matrix Estimation}
\label{sec:exp:gram}
We first illustrate that our embedding, using the parameter selections as above, can approximate the Jensen-Shanon kernel well. 
We compare three different approaches to estimating $K(p_i,p_j)=\exp(-\frac{1}{2\sigma^2}\JS(p_i,p_j))$. Each approach uses kernel density estimates $\hat{p}_i$. The estimates are compared on a dataset of $N=50$ random \acro{GMM} distributions $\{p_i\}_{i=1}^N$ and samples of size $n=2\,500$: $\chi_i = \{X_j^{(i)}\in [0,1]^2\}_{j=1}^{n} \simiid p_i$.
See the appendix for more details.

The first approach approximates \acro{JS} based on empirical estimates of entropies $\E \log \hat{p}_i$. 
The second approach estimates \acro{JS} as the Euclidean distance of vectors of projection coefficients \eqref{eq:Anorms} :
$\JS_\mathrm{pc}(p_i,p_j) = \norm{\hat{A}(\hat{p}_i) - \hat{A}(\hat{p}_j)}^2$. 
For these first two approaches we compute the pairwise kernel evaluations in the Gram matix as $G_{ij}^{\mathrm{ent}}=\exp(-\frac{1}{2\sigma^2}\JS_{\mathrm{ent}}(p_i,p_j))$, and $G_{ij}^{\mathrm{pc}}=\exp(-\frac{1}{2\sigma^2}\JS_{\mathrm{pc}}(p_i,p_j))$ using their respective approximations for \acro{JS}.
Lastly, we directly estimate the JS kernel with dot products of our random features \eqref{eq:fullapprox}:
$G_{ij}^{\mathrm{rks}} = z(\hat{A}(\hat{p}_i))\tp z(\hat{A}(\hat{p}_j))$,  with $D = 7\,000$.

%\begingroup
\setlength{\intextsep}{2mm}
\setlength{\columnsep}{5mm}%
\begin{wrapfigure}{r}{0.25\textwidth}
  \includegraphics[width=0.25\textwidth]{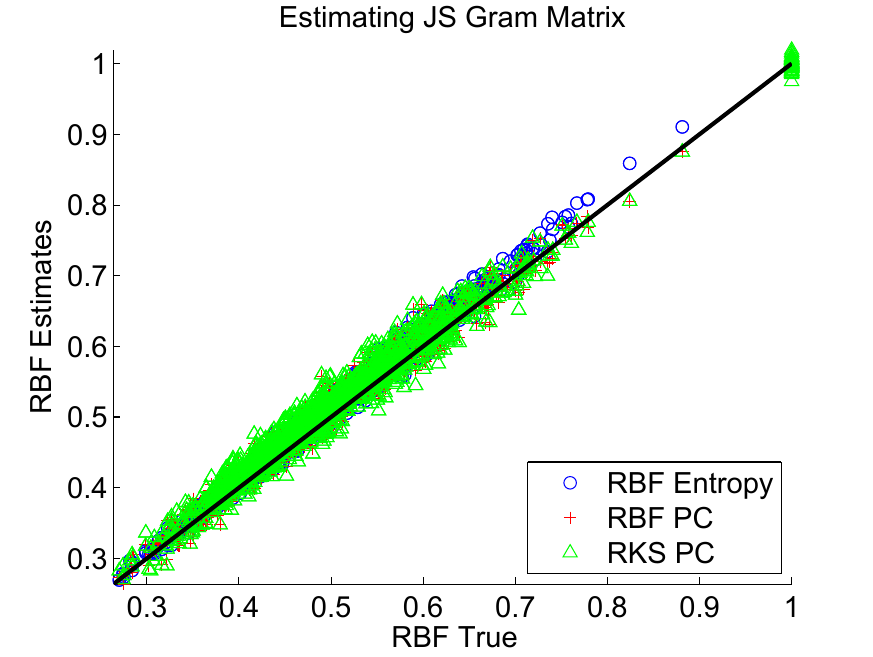}
  \caption{\label{fig:js_est} Estimating \acro{RBF} with JS divergence. }
\end{wrapfigure}
%\endgroup
\cref{fig:js_est} shows the $N^2$ true pairwise kernel values versus the aforementioned estimates. Quantitatively, the entropy method obtained a squared correlation to the true kernel value of $R_\mathrm{ent}^2 = 0.981$;
using the $A$ features with an exact kernel yielded $R_\mathrm{pc}^2 = 0.974$; 
adding \acro{rks} embeddings gave $R_\mathrm{rks}^2 = 0.966$.
Thus our method's estimates are nearly as good as direct estimation via entropies,
while allowing us to work in primal space and avoid $N \times N$ Gram matrices.

\subsection{Estimating the Number of Mixture Components} 
\label{sec:exp:mix}
\begin{figure}[!ht]
    \centering
        \includegraphics[width=.23\textwidth]{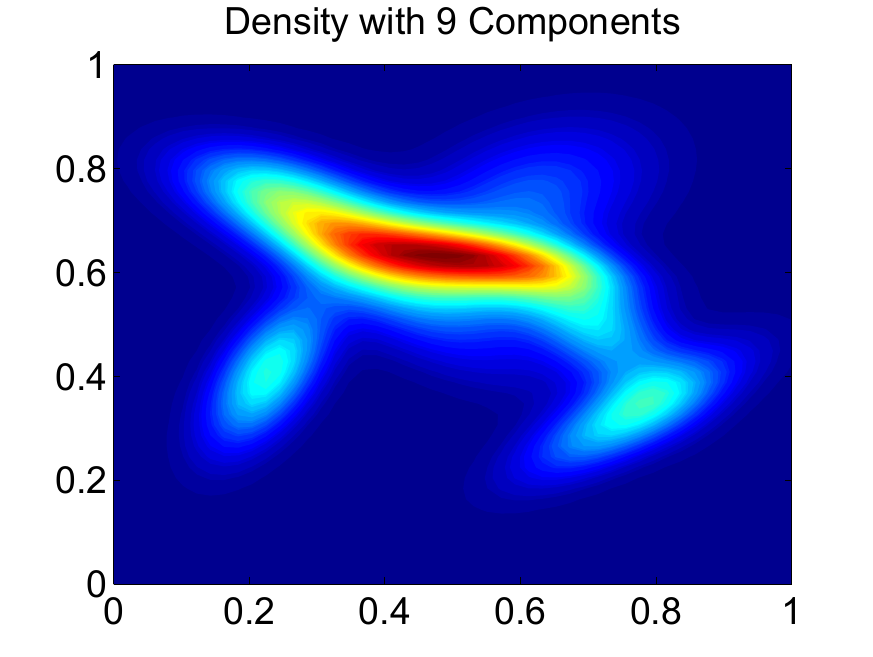}
        \includegraphics[width=.23\textwidth]{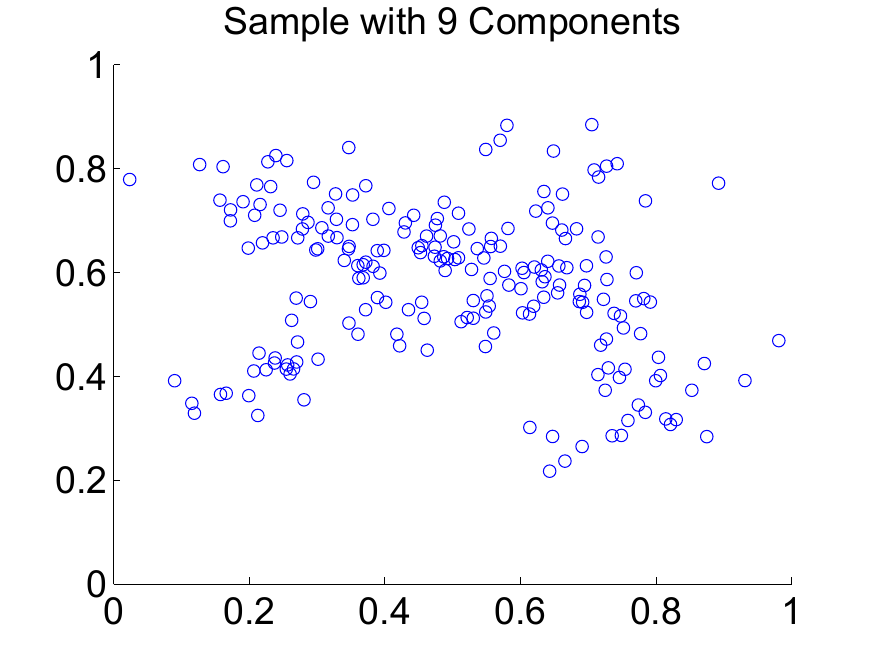}
    \caption{\label{fig:gmmEx}A \acro{GMM} and $200$ points drawn from it.}
\end{figure}

We will now illustrate the efficacy of \acro{HDD} random features in a regression task,
following \citet{Oliva2014}:
estimate the number of components from a mixture of truncated Gaussians.
We generate the distributions as follows:
Draw the number of components $Y_i$ for the $i$th distribution as $Y_i \sim \Unif\{1,\ldots,10\}$.
For each component select a mean $\mu_k^{(i)}\sim\Unif[-5,5]^2$ and covariance $\Sigma_k^{(i)} = a_k^{(i)}A_k^{(i)}A_k^{(i)\mathsf{T}}+B_k^{(i)}$,
where $a\sim\Unif[1,4]$, $A_k^{(i)}(u,v)\sim\Unif[-1,1]$, and $B_k^{(i)}$ is a diagonal $2\times2$ matrix with $B_k^{(i)}(u,u)\sim\Unif[0,1]$.
Then weight each component equally in the mixture.
Given a sample $\chi_i$, we predict the number of components $Y_i$.
An example distribution and sample are shown in \cref{fig:gmmEx};
predicting the number of components is difficult even for humans.

\begin{figure*}[!t]
  \centering
  \begin{subfigure}[b]{.48\textwidth}
      \includegraphics[width=.95\textwidth]{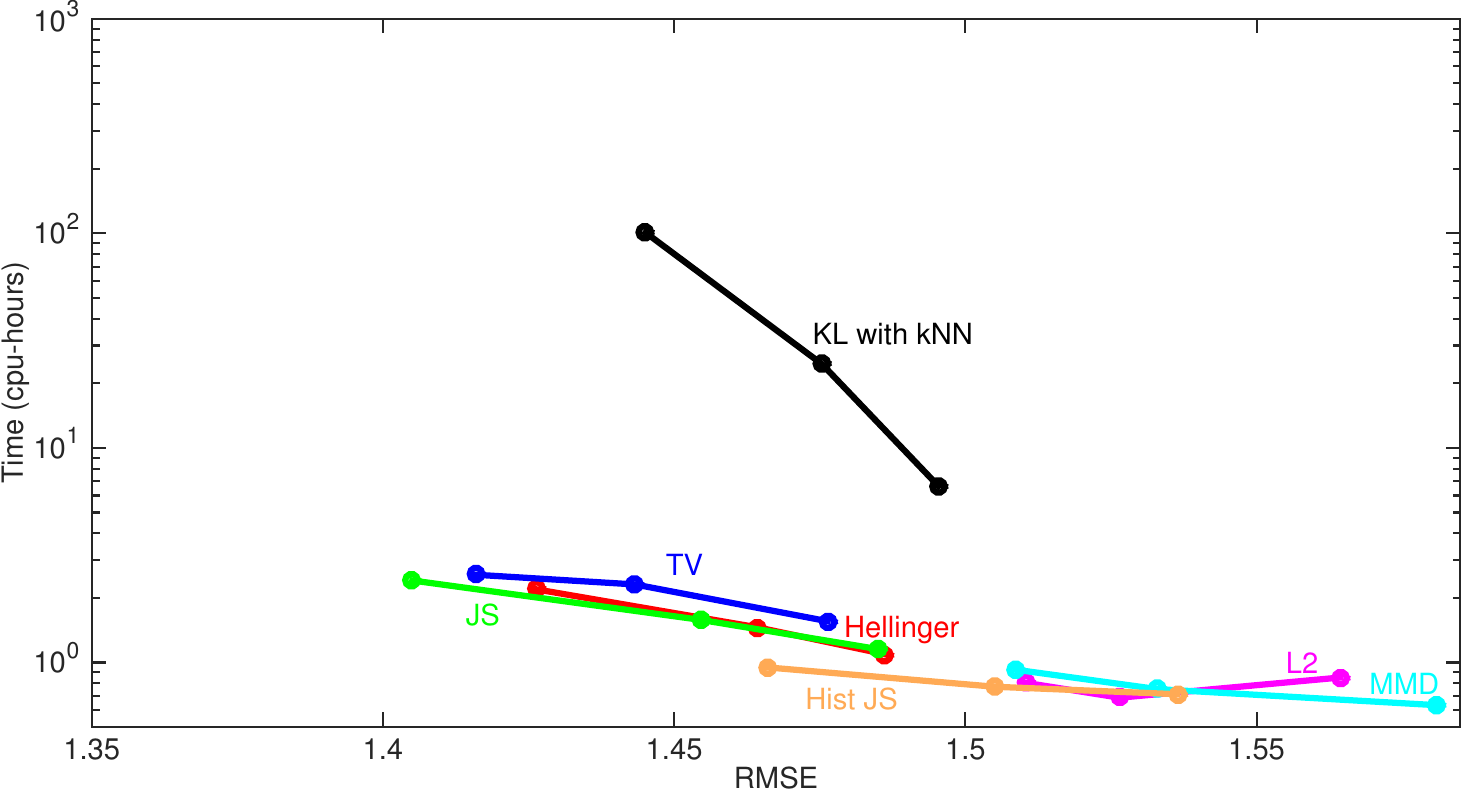}
      \caption{Samples of size 200.}
  \end{subfigure}
  ~
  \begin{subfigure}[b]{.48\textwidth}
      \includegraphics[width=.95\textwidth]{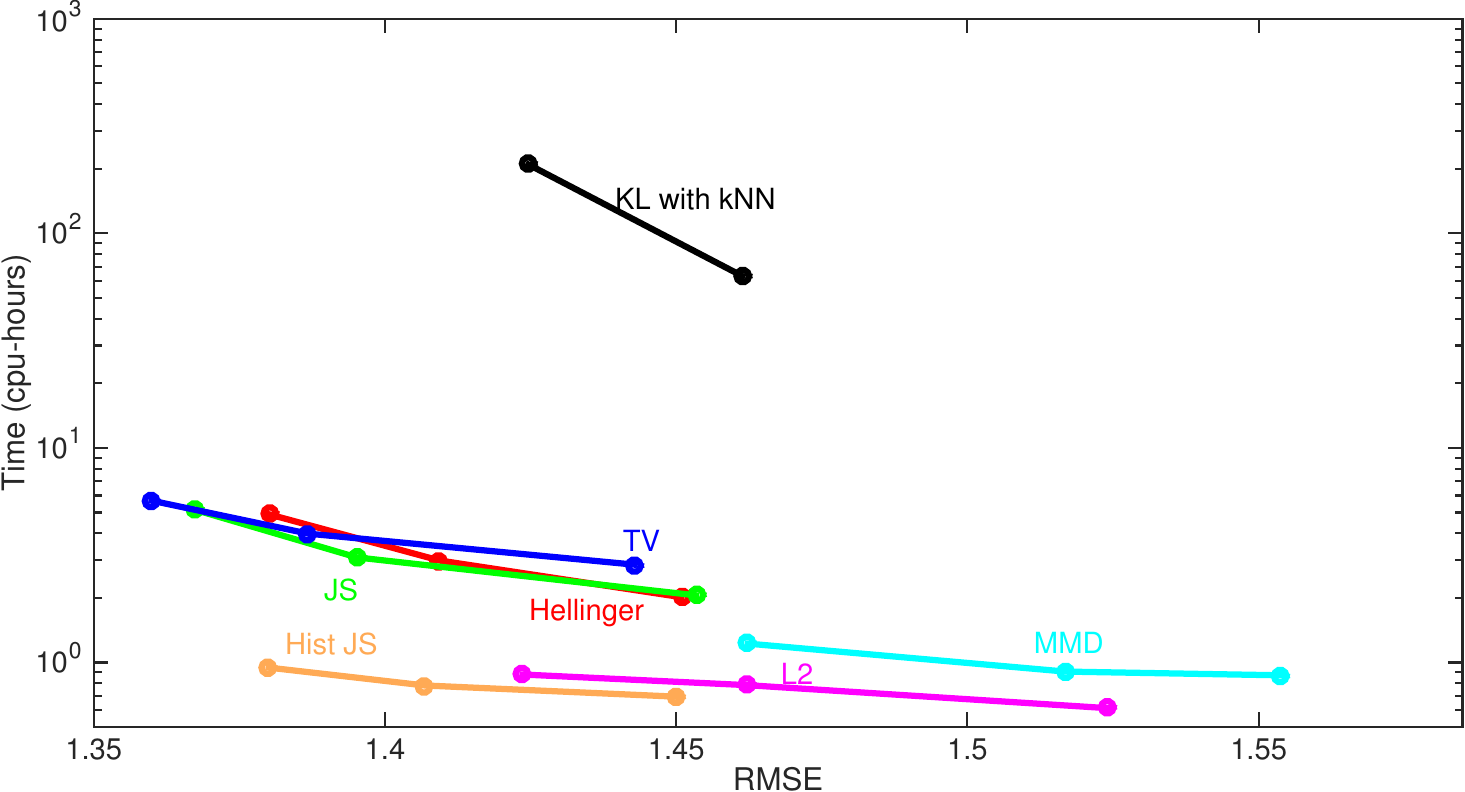}
      \caption{Samples of size 800.}
  \end{subfigure}
\caption{\label{fig:mix-res} Error and computation time for estimating the number of mixture components. The three points on each line correspond to training set sizes of 4\acro{K}, 8\acro{K}, and 16\acro{K}; error is on the fixed test set of size 2\acro{K}. Note the logarithmic scale on the time axis. The \acro{kl} kernel for $\abs{\chi_i} = 800$ with 16\acro{K} training sets was too slow to run. \acro{aic}-based predictions achieved \acro{RMSE}s of 2.7 (for 200 samples) and 2.3 (for 800); \acro{bic} errors were 3.8 and 2.7; a constant predictor of 5.5 had \acro{rmse} of 2.8.}
\end{figure*}

\begin{figure*}[!ht]
  \centering
  \begin{minipage}[b]{.49\textwidth}
    \centering
    \includegraphics[width=.9\textwidth]{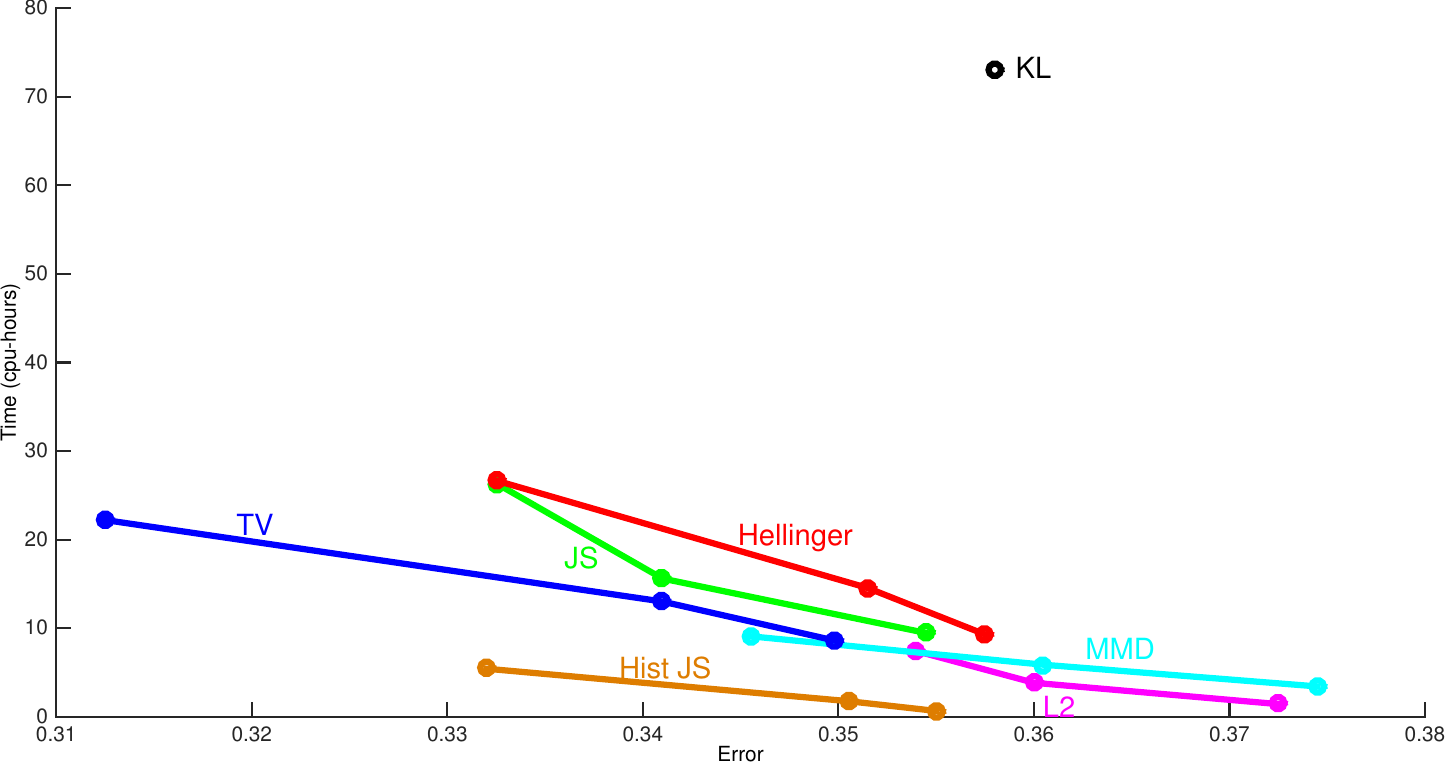}
    \caption{Misclassification rate and computation time
        for 
    classifying \acro{cifar-10} cats versus dogs.
    The three points on each line show training set sizes of 2.5\acro{k}, 5\acro{K}, and 10\acro{k}; error is on the fixed test set of size 2\acro{K}.
    Note the linear scale for time.
    The \acro{KL} kernel was too slow to run for 5\acro{K} or 10\acro{K} training points.}
    \label{fig:img-res}
  \end{minipage}
  ~
  \begin{minipage}[b]{.49\textwidth}
    \centering
    \includegraphics[width=.9\textwidth]{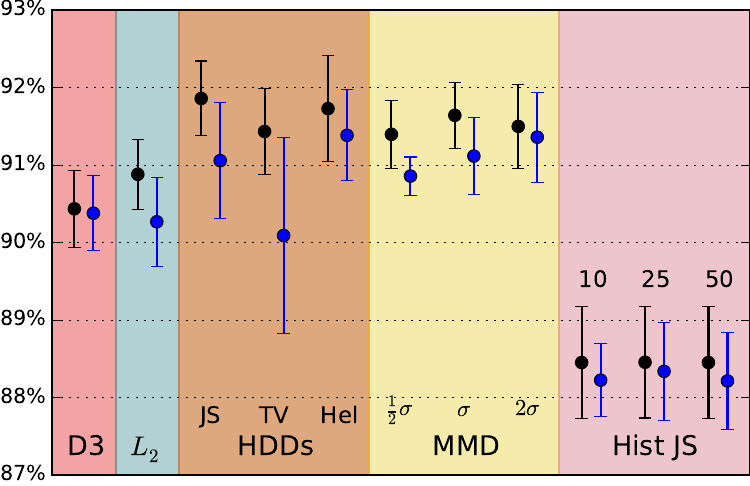}
    \caption{\label{fig:scene-res} Mean and standard deviation of accuracies on the Scene-15 dataset in 10 random splits.
    The left, black lines use $\hat{A}(\cdot)$ features; the right, blue lines show $z(\hat A(\cdot))$ features.
    \acro{MMD} methods vary bandwidth are relative to $\sigma$, the median of pairwise distances; histogram methods vary the number of bins.}
  \end{minipage}
\end{figure*}

\Cref{fig:mix-res} presents results for predicting with ridge regression the number of mixture components $Y_i$, given a varying number of sample sets $\chi_i$, with $\abs{\chi_i}\in\{200,800\}$;
we use $D = 5\,000$.
The \acro{HDD}-based kernels achieve substantially lower error than the $L_2$ and \acro{MMD} kernels.
They also outperform the histogram kernel, especially with $\abs{\chi_i} = 200$,
and the \acro{KL} kernel.
Note that fitting mixtures with \acro{EM} and selecting a number of components using \acro{AIC} \citep{aic} or \acro{BIC} \citep{bic} performed much worse than regression; only \acro{aic} with $\abs{\chi_i} = 800$ outperformed a constant predictor of $5.5$.
Linear versions of the $L_2$ and \acro{MMD} kernels were also no better than the constant predictor.

The \acro{HDD} embeddings were more computationally expensive than the other embeddings,
but much less expensive than the \acro{KL} kernel, which grows at least quadratically in the number of distributions.
Note that the histogram embeddings used an optimized \texttt{C} implementation \citep{vlfeat},
as did the \acro{KL} kernel\footnote{\scriptsize\url{https://github.com/djsutherland/skl-groups/}},
while the \acro{HDD} embeddings used a simple Matlab implementation.

\subsection{Image Classification} \label{sec:exp:img}

As another example of the performance of our embeddings,
we now attempt to classify images based on their distributions of pixel values.
We took the ``cat'' and ``dog'' classes from the \acro{CIFAR-10} dataset \citep{cifar},
and represented each $32 \times 32$ image by a set of triples $(x, y, v)$,
where $x$ and $y$ are the position of each pixel in the image
and $v$ the pixel value after converting to grayscale.
The horizontal reflection of the image was also included,
so each sample set $\chi_i \subset \R^3$ had $\abs{\chi_i} = 2\,048$.
This is certainly not the best representation for these images;
rather, we wish to show that given this simple representation,
our \acro{HDD} kernels perform well relative to the other options.

We used the same kernels as above
in an \acro{svm} classifier from \acro{liblinear} (\citeauthor{liblinear}~\citeyear{liblinear}, for the embeddings)
or \acro{libsvm} (\citeauthor{libsvm}~\citeyear{libsvm}, for the \acro{kl} kernel),
with $D = 7\,000$.
\Cref{fig:img-res} shows computation time and accuracy on the standard test set (of size 2\acro{K})
with 2.5\acro{K}, 5\acro{K}, and 10\acro{K} training images.
Our \acro{JS} and Hellinger embedding approximately match the histogram \acro{JS} embedding in accuracy here,
while our \acro{TV} embedding beats histogram \acro{JS};
all outperform $L_2$ and \acro{MMD}.
We could only run the \acro{KL} kernel for the 2.5\acro{K} training set size;
its accuracy was comparable to the \acro{HDD} and histogram embeddings,
at far higher computational cost.

\subsection{Scene Classification} \label{sec:exp:scene}
Modern computer vision classification systems typically consist of a deep network with several convolutional and pooling layers to extract complex features of input images,
followed by one or two fully-connected classification layers.
The activations are of shape $n \times h \times w$,
where $n$ is the number of filters;
each unit corresponds to an overlapping patch of the original image.
We can thus treat the final pooled activations as a sample of size $h w$ from an $n$-dimensional distribution,
similarly to how \citet{poczos2012nonparametric} and \citet{muandet2012learning} used \acro{sift} features from image patches.
\citet{d3} set accuracy records on several scene classification datasets with a particular ad-hoc method of extracting features from distributions ({\small \textsf{D3}});
we compare to our more principled alternatives.

We consider the Scene-15 dataset \citep{scene-15},
which contains $4\,485$ natural images in 15 location categories,
and follow \citeauthor{d3} in extracting features from the last convolutional layer of the \texttt{imagenet-vgg-verydeep-16} model \citep{verydeep}.
We replace that layer's rectified linear activations
with sigmoid squashing to $[0, 1]$.\footnote{
We used piecewise-linear weights before the sigmoid function such that $0$ maps to $0.5$, the 90th percentile of the positive observations maps to $0.9$, and the 10th percentile of the negative observations to $0.1$, for each filter.
}
$hw$ ranges from $400$ to $1\,000$.
There are 512 filter dimensions;
we concatenate features $\hat{A}(\hat{p}_i)$ extracted from each independently.

We train on the standard for this dataset of 100 images from each class (1500 total) and test on the remainder;
\cref{fig:scene-res} shows results.
We do not add any spatial information to the model;
still, we match the best prior published performance of $91.59 \pm 0.48$, which trained on over 7 million external images \citep{places}.
Adding spatial information brought the D3 method slightly above 92\% accuracy;
their best hybrid method obtained 92.9\%.
Using these features, however,
our methods match or beat \acro{MMD}
and substantially outperform D3, $L_2$, and the histogram embeddings.

\section{Discussion}
This work presents the first nonlinear embedding of density functions for quickly computing \acro{HDD}-based kernels, including kernels based on the popular total variation, Hellinger and Jensen-Shanon divergences. 
While such divergences have shown good empirical results in the comparison of densities, nonparametric uses of kernels with these divergences previously necessitated the computation of a large $N\times N$ Gram matrix,
prohibiting their use in large datasets.
Our embeddings allow one to work in a primal space while using information theoretic kernels.
We analyze the approximation error of our embeddings, and illustrate their quality on several synthetic and real-world datasets.

\clearpage
\fontsize{9.5pt}{10.5pt} \selectfont
\bibliographystyle{aaai}
\bibliography{paper}

\clearpage
\section{Acknowledgements}
This work was funded in part by NSF grant IIS1247658 and by DARPA grant FA87501220324.
DJS is also supported by a Sandia Campus Executive Program fellowship.

\appendix
\section{Appendices}

\subsection{Gram Matrix Estimation}
We illustrate our embedding's ability to approximate the Jensen-Shanon divergence. In the examples below the densities considered are mixtures of five equally weighted truncated spherical Gaussians on $[0,1]^2$. That is, 
\begin{align*}
p_i(x) = \frac{1}{5}\sum_{j=1}^5 \N_t(m_{ij}, \diag(s^2_{ij}))
\end{align*}
where $m_{ij} \simiid \Unif([0,1]^2)$, $s_{ij} \simiid \Unif([ 0.05, 0.15]^2)$
and $\N_t(m,s)$ is the distribution of a Gaussian truncated on $[0,1]^2$ with mean parameter $m$ and covariance matrix parameter $s$. We work over the sample set $\{\chi_i\}_{i=1}^N$, where $\chi_i = \{X_j^{(i)}\in [0,1]^2\}_{j=1}^{n} \simiid p_i$, $n=2500$, $N=50$.

We compare three different approaches to estimating $K(p_i,p_j)=\exp(-\frac{1}{2\sigma^2}\JS(p_i,p_j))$. Each approach uses density estimates $\hat{p}_i$, which are computed using kernel density estimation. The first approach is based on estimating \acro{JS} using empirical estimates of entropies:
\begin{align*}
&\JS(p_i,p_j) \\
&= -\tfrac{1}{2}\E_{p_i}\left[\log\left(\frac{1}{p_i(x)}\right)\right] -\tfrac{1}{2}\E_{p_j}\left[\log\left(\frac{1}{p_j(x)}\right)\right] \\
&\quad+ \E_{\frac{1}{2}p_i+\frac{1}{2}p_j}\left[\log\left(\frac{2}{p_i(x)+p_j(x)}\right)\right]\\
&\approx -\tfrac{1}{2}\sum_{m=1}^{\left \lceil{n/2}\right \rceil }\log\left( \frac{1}{\hat{p}_i(X_m^{(i)})} \right)
-\tfrac{1}{2}\sum_{m=1}^{\left \lceil{n/2}\right \rceil }\log\left(\frac{1}{\hat{p}_j(X_m^{(j)})}\right) \\
&\quad+\tfrac{1}{2}\sum_{m=1}^{\left \lceil{n/2}\right \rceil }\log\left(\frac{2}{\hat{p}_i(X_m^{(i)})+\hat{p}_j(X_m^{(i)})}\right) \\
&\quad+\tfrac{1}{2}\sum_{m=1}^{\left \lceil{n/2}\right\rceil }\log\left(\frac{2}{\hat{p}_i(X_m^{(j)})+\hat{p}_j(X_m^{(k)})}\right)\\
&= \JS_{\mathrm{ent}}(p_i,p_j),
\end{align*}
where density estimates $\hat{p_i}$ above are based on points $\{X_m^{(i)}\}_{m=\left\lceil{n/2}\right\rceil+1}^{n}$ to avoid biasing the empirical means.
The second approach estimates \acro{JS} as the Euclidean distance of vectors of projection coefficients:
\begin{align*}
\JS(p_i,p_j) &\approx \norm{\hat{A}(\hat{p}_i) - \hat{A}(\hat{p}_j)}^2 =\JS_{\mathrm{pc}}(p_i,p_j),
\end{align*}
where here the density estimates $\hat{p_i}$ are based on the entire set of points $\chi_i$.
We build a Gram matrix for each of these approaches by
setting $G_{ij}^{\mathrm{ent}}=\exp(-\frac{1}{2\sigma^2}\JS_{\mathrm{ent}}(p_i,p_j))$ and $G_{ij}^{\mathrm{pc}}=\exp(-\frac{1}{2\sigma^2}\JS_{\mathrm{pc}}(p_i,p_j))$.
Lastly, we directly estimate the \acro{JS} kernel with random features:
\begin{align*}
G_{ij}^{\mathrm{rks}} = z(\hat{A}(\hat{p}_i))\tp z(\hat{A}(\hat{p}_j)).
\end{align*}

\begin{figure}[t]
  \centering
  \includegraphics[width=.5\textwidth]{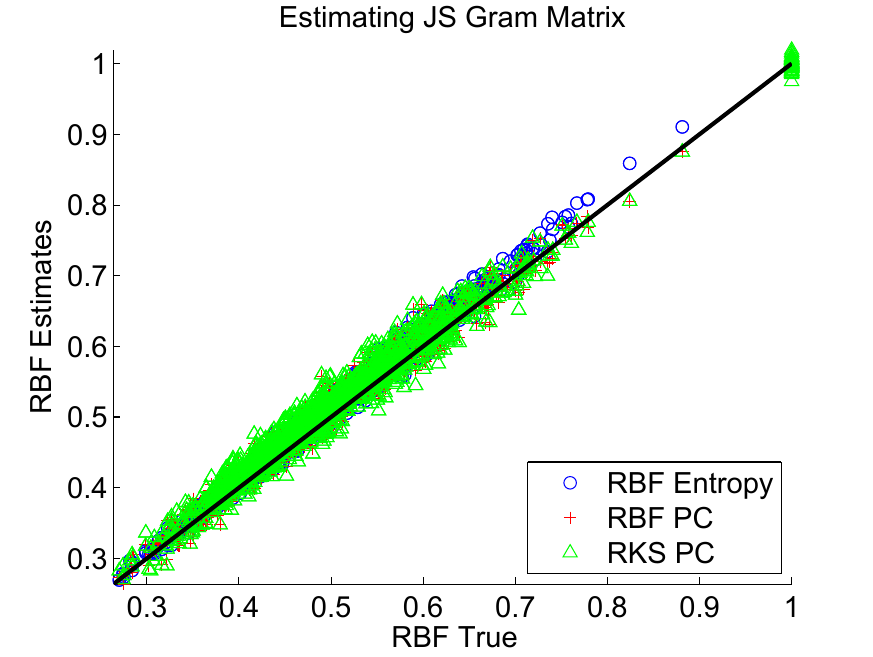}
    \caption{\label{fig:js_est_large}Estimating \textsc{RBF} values using the \textsc{JS} diveregence.}
\end{figure}

We compare the effectiveness of each approach by computing the $R^2$ score of the estimates produced versus a true \acro{JS} kernel value computed through numerically integrating the true densities (see \cref{fig:js_est_large,tbl:js_r2}). The \acro{RBF} values estimated with our random features produce estimates that are nearly as good as directly estimating JS divergences through entropies, whilst allowing us to work over a primal space and thus avoid computing a $N\times N$ Gram matrix for learning tasks.
\begin{table}[!ht]
\caption{\label{tbl:js_r2}$R^2$ values of estimates of JS Gram elements. \vspace{1mm}}
\centering
\begin{tabular}{l{c}}
Method      & $R^2$\\
\hline
Entropies   & 0.9812  \\
\acro{PC}s  & 0.9735  \\
\acro{RKS}  & 0.9662  \\
\end{tabular}
\end{table}

\subsection{Proofs}

We will now prove the bound on the error probability of our embedding
$\Pr\left(\Abs{K(p, q) - z(\hat{A}(\hat{p}))\tp z(\hat{A}(\hat{q}))} \ge \varepsilon\right)$
for fixed densities $p$ and $q$.

\paragraph{Setup}
We will need a few assumptions on the densities:
\begin{enumerate}
\item $p$ and $q$ are bounded above and below: for $x \in [0, 1]^\ell$, $0 < \rho_* \le p(x), q(x) \le \rho^* < \infty$.
    \label{assump:bounded}
\item $p, q \in \Sigma(\beta, L_\beta)$ for some $\beta, L_\beta > 0$.
    $\Sigma(\beta, L)$ refers to the H\"older class of functions $f$ whose
    partial derivatives up to order $\floor\beta$ are continuous
    and whose $r$th partial derivatives, where $r$ is a multi-index of order $\floor\beta$,
    satisfy $\Abs{D^r f(x) - D^r f(y)} \le L \norm{x - y}^\beta$.
    Here $\floor \beta$ is the greatest integer \textit{strictly} less than $\beta$.
    \label{assump:holder}
\item $p$, $q$ are periodic.
    \label{assump:periodic}
\end{enumerate}
These are fairly standard smoothness assumptions in the nonparametric estimation
literature.

Let $\gamma = \min(\beta, 1)$.
If $\beta > 1$, then $p, q \in \Sigma(1, L_\gamma)$ for some $L_\gamma$;
otherwise, clearly $p, q \in \Sigma(\beta, L_\beta)$.
Then, from assumption \ref{assump:periodic},
$p, q \in \Sigma_\text{per}(\gamma, L_\gamma)$,
the periodic H\"older class.
We'll need this to establish the Sobolev ellipsoid containing $p$ and $q$.

We will use kernel density estimation with a
bounded, continuous kernel so that the bound of \citet{Gine2002} applies,
with bandwidth $h \asymp n^{-1/(2 \beta + \ell)} \log n$,
and truncating density estimates to $[\rho_*, \rho^*]$.

We also use the Fourier basis $\varphi_\alpha = \exp\left( 2 \ii \pi \alpha\tp x \right)$,
and define $V$ as the set of indices $\alpha$ s.t.
$\sum_{j=1}^\ell \abs{\alpha_j}^{2 s} \le t$ for parameters $0 < s \le 1$, $t > 0$ to be discussed later.

\paragraph{Decomposition}
Let $r_{\sigma}(\Delta) = \exp\left( - \Delta^2 / (2 \sigma^2) \right)$.
Then
\begin{multline*}
  \Abs{K(p, q)
     - z(\hat{A}(\hat{p}))\tp \, z(\hat{A}(\hat{q}))}
  \le
\\  \Abs{K(p, q)
     - r_{\sigma_k}\left( \norm{\hat{A}(\hat{p}) - \hat{A}(\hat{q})} \right)}
\\  +
  \Abs{r_{\sigma_k}\left( \norm{\hat{A}(\hat{p}) - \hat{A}(\hat{q})} \right)
     - z(\hat{A}(\hat{p}))\tp \, z(\hat{A}(\hat{q}))}
.\end{multline*}
The latter term was bounded by \citet{rahimi2007random}.
For the former, note that $r_\sigma$ is $\frac{1}{\sigma \sqrt{e}}$-Lipschitz,
so the first term is at most $\frac{1}{\sigma_k \sqrt{e}} \Abs{d(p, q) - \norm{\hat{A}(\hat{p}) - \hat{A}(\hat{q})}}$.
Breaking this up with the triangle inequality:
\begin{multline}
    \Abs{d(p, q) - \norm{\hat{A}(\hat{p}) - \hat{A}(\hat{q})}}
\le\\
\Abs{d(p, q) - d(\hat{p}, \hat{q})}
+ \Abs{d(\hat{p}, \hat{q})
- \norm{\psi(\hat{p}) - \psi(\hat{q})}}
\\+ \Abs{\norm{\psi(\hat{p}) - \psi(\hat{q})} - \norm{A(\hat{p}) - A(\hat{q})}} \\
+ \Abs{\norm{A(\hat{p}) - A(\hat{q})} - \norm{\hat{A}(\hat{p}) - \hat{A}(\hat{q})}}
\label{eq:big-decomposition}
.\end{multline}

\paragraph{Estimation error}
Recall that $d$ is a metric, so the reverse triangle inequality allows us to address the first term with
\[
    \Abs{d(p, q) - d(\hat{p}, \hat{q})}
    \le d(p, \hat{p}) + d(q, \hat{q})
.\]
For $d^2$ the total variation, squared Hellinger, or Jensen-Shannon \acro{HDD}s,
we have that $d^2(p, \hat{q}) \le \TV(p, \hat{p})$ \citep{lin1991divergence}.
Moreover, as the distributions are supported on $[0, 1]^\ell$,
$\TV(p, \hat{p}) = \tfrac12 \Norm{p - \hat{p}}_1 \le \tfrac12 \Norm{p - \hat{p}}_\infty$.

It is a consequence of \citet{Gine2002} that, for any $\delta > 0$,
$\Pr\left( \Norm{p - \hat{p}}_\infty > \frac{\sqrt{C_\delta \log n}}{n^{\beta / (2 \beta + \ell)}} \right) < \delta$
for some $C_\delta$ depending on the kernel.
Thus $\Pr\left( \Abs{d(p, q) -d(\hat{p}, \hat{q})} \ge \varepsilon \right)
      < 2 C^{-1}\left( \frac{\varepsilon^4 n^{2 \beta / (2 \beta + \ell)}}{4 \log n} \right)$,
where $C_{C^{-1}(x)} = x$.

\paragraph{$\lambda$ approximation}
The second term of \eqref{eq:big-decomposition},
the approximation error due to sampling $\lambda$s,
admits a simple Hoeffding bound.
Note that $\Norm{\hat{p}_\lambda^R - \hat{q}_\lambda^R}^2 + \Norm{\hat{p}_\lambda^I - \hat{q}_\lambda^I}^2$,
viewed as a random variable in $\lambda$ only,
has expectation $d^2(\hat{p}, \hat{q})$
and is bounded by $[0, 4Z]$ (where $Z = \int_{\R_{\ge 0}} \ud\mu(\lambda)$):
write it as $Z \int \abs{\hat{p}(x)^{\tfrac12 + \ii\lambda} - \hat{q}(x)^{\tfrac12 + \ii\lambda}}^2 \,\ud x$,
expand the square, and use $\int \sqrt{\hat{p}(x) \hat{q}(x)} \ud x \le 1$ (via Cauchy-Schwarz).

For nonnegative random variables $X$ and $Y$,
$\Pr\left( \Abs{X - Y} \ge \varepsilon \right) \le \Pr\left( \Abs{X^2 - Y^2} \ge \varepsilon^2 \right)$,
so we have that
$\Pr\left( \Abs{\Norm{\psi(\hat{p}) - \psi(\hat{q})} - d(\hat{p}, \hat{q})} \ge \varepsilon \right)$ is at most $2 \exp(- M \varepsilon^4 / (8 Z^2))$.

\paragraph{Tail truncation error}
The third term of \eqref{eq:big-decomposition},
the error due to truncating the tail projection coefficients of the $p_\lambda^S$ functions,
requires a little more machinery.
First note that $\Abs{\Norm{\psi(\hat{p}) - \psi(\hat{q})}^2 - \Norm{A(\hat{p}) - A(\hat{q})}^2}$ is at most
\begin{equation}
    \sum_{j=1}^M \sum_{S=R,I} \sum_{\alpha \notin V} \Abs{a_{\alpha}(\hat{p}_\lambda^S - \hat{q}_\lambda^S)}^2
.\end{equation}

Let $\mathcal{W}(s, L)$ be the Sobolev ellipsoid of functions $\sum_{\alpha \in \Z^\ell} a_\alpha \varphi_\alpha$ such that $\sum_{\alpha \in \Z^\ell} \left( \sum_{j=1}^\ell \abs{\alpha_j}^{2 s} \right) \abs{a_\alpha}^2 \le L$,
where $\varphi$ is still the Fourier basis.
Then Lemma 14 of \citet{renyi-friends} shows that
$\Sigma_\text{per}(\gamma, L_\gamma) \subseteq \mathcal{W}(s, L')$
for any $0 < s < \gamma$
and $L' = \ell L_\gamma^2 (2 \pi)^{-2 \floor{\gamma}} \frac{4^\gamma}{4^\gamma - 4^s}$.

So, suppose that $\hat{p}, \hat{q} \in \Sigma_\text{per}(\hat\gamma, \widehat{L})$ with probability at least $1 - \delta$.
Since $x \mapsto x^{\tfrac12 + \ii\lambda}$ is $\frac{\sqrt{1 + 4 \lambda^2}}{2 \sqrt{\rho_*}}$-Lipschitz on $[\rho_*, \infty)$, 
$\hat{p}_\lambda^S \in \Sigma_\text{per}\left(\hat\gamma, \tfrac12 \sqrt{1 + 4\lambda^2} \, \widehat{L} \, \rho_*^{-\tfrac12} \right)$
and so
%$\hat{p}_\lambda^S - \hat{q}_\lambda^S \in \Sigma_\text{per}\left(\hat\gamma, \sqrt{1 + 4\lambda^2} \, \widehat{L} \, \rho_*^{-\tfrac12}\right)$
$\hat{p}_\lambda^S - \hat{q}_\lambda^S$ is in $\mathcal{W}(s, (1 + 4 \lambda^2) \widehat{L}')$
for $s < \hat\gamma$
and $\widehat{L}' = \ell \widehat{L}^2 \rho_*^{-1} / (1 - 4^{s-\hat\gamma})$.

Recall that we chose $V$ to be the set of $\alpha \in \Z^\ell$ such that $\sum_{j=1}^\ell \abs{\alpha_j}^{2 s} \le t$.
Thus $\sum_{\alpha \notin V} \abs{a_\alpha(\hat{p}_\lambda^S - \hat{q}_\lambda^S)}^2
\le  \sum_{\alpha \notin V} \abs{a_\alpha(\hat{p}_\lambda^S - \hat{q}_\lambda^S)}^2 \left( \sum_{j=1}^\ell \abs{\alpha_j}^{2 s} \right) / t
\le (1 + 4 \lambda^2) \widehat{L}' / t$.

The tail error term is therefore at least $\varepsilon$ with probability no more than
$
\delta
+ 2 \sum_{j=1}^M
    \Pr\left( (1 + 4 \lambda_j^2) \widehat{L}' / t \ge \varepsilon^2 / (2 M) \right)
$.
The latter probability, of course, depends on the choice of \acro{HDD} $d$.
Letting $\zeta = t \varepsilon^2/(8 M \widehat{L}') - \tfrac14$,
it is 1 if $\zeta < 0$ and $1 - \mu\left( [0, \sqrt\zeta] \right) / Z$ otherwise.
If $\zeta \ge 0$,
squared Hellinger's probability is 0,
and total variation's is $\frac{2}{\pi} \arctan(\sqrt\zeta)$.
A closed form for the cumulative distribution function for the Jensen-Shannon measure is unfortunately unknown.

\paragraph{Numerical integration error}
The final term of \eqref{eq:big-decomposition} also bears a Hoeffding bound.
Define the projection coefficient difference
$\Delta^S_{\lambda,\alpha}(p, q) = a_{\alpha,\lambda}(p_\lambda^S) - a_\alpha(q_\lambda^S)$,
and $\hat\Delta$ similarly but with $\hat{a}$.
Then $\Abs{\Norm{A(\hat{p}) - A(\hat{q})}^2 - \Norm{\hat{A}(\hat{p}) - \hat{A}(\hat{q})}^2}$
is at most
\begin{equation}
\sum_{j=1}^M \sum_{S = R, I} \sum_{\alpha \in V}
\Abs{
    \Abs{\Delta^S_{\alpha,{\lambda_j}}(\hat{p}, \hat{q})}^2
  - \Abs{\hat{\Delta}^S_{\alpha,{\lambda_j}}(\hat{p}, \hat{q})}^2
}
\label{eq:numint}
.\end{equation}
Letting $\hat\epsilon(p) = a_\alpha(\hat{p}_\lambda^S) - \hat{a}_\alpha(\hat{p}_\lambda^S)$,
each summand is at most $(\hat\epsilon(p) + \hat\epsilon(q))^2 + 2 \Abs{\Delta^S_{\lambda,\alpha}(\hat{p}, \hat{q})} (\hat\epsilon(p) + \hat\epsilon(q))$.
Also, $\Abs{\Delta^S_{\alpha,\lambda}(\hat{p}, \hat{q})} \le 2 \sqrt{Z}$,
using Cauchy-Schwarz on the integral and $\Norm{\varphi_\alpha}_2 = 1$.
Thus each summand in \eqref{eq:numint}
can be more than $\varepsilon$ only if one of the $\hat\epsilon$s is more than $\sqrt{Z + \varepsilon/4} - \sqrt{Z}$.

Now, using \eqref{eq:ApNumInt},
$\hat{a}_\alpha(\hat{p}_{\lambda}^S)$ is an empirical mean of $n_e$ independent terms,
each with absolute value bounded by $(\sqrt{\rho^*} + 1) \max_x \abs{\varphi_\alpha(x)} = \sqrt{\rho^*} + 1$.
Thus, using a Hoeffding bound on the $\hat\epsilon$s, we get that
$\Pr\left( \Abs{\norm{A(\hat{p}) - A(\hat{q})}^2 - \norm{\hat{A}(\hat{p}) - \hat{A}(\hat{q})}^2} \ge \varepsilon \right)
$ is no more than $8 M S \exp\left( - \frac{n_e \left( \sqrt{Z + \varepsilon^2 / (8 S)} - \sqrt{Z} \right)^2}{ 2 Z (\sqrt{\rho^*} + 1)^2 } \right)$.

\paragraph{Final bound}
Combining the bounds for the decomposition \eqref{eq:big-decomposition} with the pointwise rate for \acro{RKS} features,
we get:
\begin{multline}
    \Pr\left( \Abs{K(p, q) - z(\hat{A}(\hat{p})\tp z(\hat{A}(\hat{q}))} \ge \varepsilon \right)
    \le
\\
      2 \exp\left( - D \varepsilon_\text{RKS}^2 \right)
    + 2 C^{-1}\left( \frac{\varepsilon^4_\text{KDE} n^{2 \beta / (2 \beta + \ell)}}{4 \log n} \right)
\\  + 2 \exp \left( - M \varepsilon_{\lambda}^4 / (8 Z^2) \right)
\\    + \delta + 2 M \left(
    1
    - \mu \left[
    0,
    \sqrt{\max\left(
        0,
        \frac{\rho_* t \varepsilon_\text{tail}^2}{8 M \ell \widehat{L}^2}
        \frac{4^{\hat\gamma} - 4^s}{4^{\hat\gamma}}
        - \frac{1}{4}
    \right)}
    \right) \right)
\\  + 8 M \,\abs{V} \exp\left( - \tfrac12 n_e \left( \frac{\sqrt{1 + \varepsilon_\text{int}^2 / (8 \,\abs{V}\, Z)} - 1}{\sqrt{\rho^*} + 1} \right)^2 \right)
\end{multline}
for any $\varepsilon_\text{RKS} + \frac{1}{\sigma_k \sqrt{e}} \left( \varepsilon_\text{KDE} + \varepsilon_\lambda + \varepsilon_\text{tail} + \varepsilon_\text{int} \right) \le \varepsilon$.

\end{document}